\documentclass[sigconf]{acmart}
\usepackage{multirow}
\usepackage{framed}
\usepackage{balance}
\AtBeginDocument{%
  \providecommand\BibTeX{{%
    \normalfont B\kern-0.5em{\scshape i\kern-0.25em b}\kern-0.8em\TeX}}}

\copyrightyear{2022}
\acmYear{2022}
\setcopyright{acmcopyright}
\acmConference[ASE]{IEEE/ACM International Conference on Automated Software Engineering}{ 2022}{}
\acmBooktitle{IEEE/ACM International Conference on Automated Software Engineering}

\begin{document}

\title{SmOOD: Smoothness-based Out-of-Distribution Detection Approach for Surrogate Neural Networks in Aircraft Design}

\author{Houssem Ben Braiek}
\affiliation{
  \institution{Polytechnique Montreal}
  \city{Montreal}
  \state{Quebec}
  \country{Canada}
  }
\email{houssem.ben-braiek@polymtl.ca}

\author{Ali Tfaily}
\affiliation{
  \institution{Polytechnique Montreal}
  \city{Montreal}
  \state{Quebec}
  \country{Canada}
  }
\email{ali.tfaily@aero.bombardier.com}

\author{Foutse Khomh}
\affiliation{
 \institution{Polytechnique Montreal}
 \city{Montreal}
  \state{Quebec}
 \country{Canada}
 }
\email{foutse.khomh@polymtl.ca}

\author{Thomas Reid}
\affiliation{
  \institution{Sycodal Inc.}
  \city{Montreal}
  \state{Quebec}
  \country{Canada}
  }
\email{t.reid@sycodal.ca}

\author{Ciro Guida}
\affiliation{
  \institution{Bombardier Aerospace Inc.}
  \city{Montreal}
  \state{Quebec}
  \country{Canada}
  }
\email{ciro.guida@aero.bombardier.com}
\renewcommand{\shortauthors}{Ben Braiek, et al.}
\newcommand{\metadata}[1]{\textcolor{blue}{#1}}
\newcommand{\name}[1]{\textit{SmOOD}}
\begin{abstract}
Aircraft industry is constantly striving for more efficient design optimization methods in terms of human efforts, computation time, and resources consumption. Hybrid surrogate optimization maintains high results quality while providing rapid design assessments when both the surrogate model and the switch mechanism for eventually transitioning to the HF model are calibrated properly. Feedforward neural networks (FNNs) can capture highly nonlinear input-output mappings, yielding efficient surrogates for aircraft performance factors. However, FNNs often fail to generalize over the out-of-distribution (OOD) samples, which hinders their adoption in critical aircraft design optimization. Through SmOOD, our smoothness-based out-of-distribution detection approach, we propose to codesign a model-dependent OOD indicator with the optimized FNN surrogate, to produce a trustworthy surrogate model with selective but credible predictions. Unlike conventional uncertainty-grounded methods, SmOOD exploits inherent smoothness properties of the HF simulations to effectively expose OODs through revealing their suspicious sensitivities, thereby avoiding over-confident uncertainty estimates on OOD samples. By using SmOOD, only high-risk OOD inputs are forwarded to the HF model for re-evaluation, leading to more accurate results at a low overhead cost. Three aircraft performance models are investigated. Results show that FNN-based surrogates outperform their Gaussian Process counterparts in terms of predictive performance. Moreover, \name{} does cover averagely $85\%$ of actual OODs on all the study cases. When SmOOD plus FNN surrogates are deployed in hybrid surrogate optimization settings, they result in a decrease error rate of $34.65\%$ and a computational speed up rate of $58.36\times$, respectively.
\end{abstract}



\maketitle

\section{Introduction}
The design of aircraft, from conceptualization to commercialization, is expensive. Their lead-time engineering process
can take many years and cost billions of dollars from design to manufacture. Multidisciplinary design optimization (MDO) leverages advances in numerical simulation and computer aided design to revolutionize aircraft design~\cite{pi2013cfd}. By providing simulation-driven design configurations at an earlier stage~\cite{pardessus2004concurrent}, next-generation aircraft will have reduced development costs and lead times. Indeed, MDO optimizes different types of objective functions~\cite{audet2016blackbox} through repetitive assessment of evolving configurations via the use of high-fidelity (HF) mathematical models, which can lead to high computational cost. Even though HF models offer a less expensive alternative to physical testing, HF simulations of aircraft flight dynamics typically take several days to assess thousands of design configurations. Various solutions have been proposed to circumvent the compute-intensive issues of MDO through surrogate modeling~\cite{sobester2008engineering}. Indeed, surrogate models, also known as metamodels, speed up the optimization process since they assess configurations approximately, but at a much lower expense and faster rate. In the context of aircraft design, developing surrogate models for aircraft design poses a challenge due to the high nonlinearity of some of the aircraft design variables of concern. Due to its accurate predictions and fast inferences, deep learning networks can effectively address the imperfect surrogate modeling issue faced by conventional machine learning models in the field of aircraft design optimization~\cite{sun2019review}. Neural networks~\cite{hornik1989multilayer} are universal approximators with non-convex learning algorithms that train fast and capture high-nonlinearity input-output mappings. Nevertheless, they fundamentally inherit a closed-world assumption~\cite{he2015delving}, and provide no guidance on how to handle out-of-distribution data (OOD)~\cite{yang2021generalized}. Indeed, their generalization capability is guaranteed on novel configurations, but drawn from the same or close distribution as the training set, called in-distribution data (ID). According to the core notions of model over-parametrization and complexity, modern neural networks can behave worse than their simpler counterparts when exposed to OOD inputs. The resulting corner-case behaviors adversely affect the overall performance assessments for the design configurations, which in turn increases the aircraft design lead time and development budget. However, chasing complete training sets that cover all the facets of the distributions in relation to the quantities of interest, restarts the onset limitation of compute-intensive costs, since the data in aircraft design is typically generated from expensive numerical simulations. Furthermore, the construction of advanced surrogate models using state-of-the-art deep learning techniques~\cite{sun2019review, cao2020dnn, du2022airfoil} requires model engineering efforts, draws on new expertise, and often results in over-optimized models that target individually specific sub-problems.\\
In response to the aformentioned challenges
, we propose \name{}, a smoothness-based OOD detection approach, that allows to codesign the surrogate neural network model for accurate assessments and its OOD detector for selective prediction. Unlike common research on out-of-distribution, \name{} exploits the apriori smoothness property of the simulated system to overcome the main challenges of OOD detection, including complex, high-dimensional input spaces and degeneration of uncertainty estimates beyond the ID regions.\\
This research work makes the following contributions:
\begin{itemize}
   \item[--] FNNs are investigated as an alternative to GPs in MDO for aircraft design at early stages.
    \item[--] Pointwise sensitivity profiles are proposed, and their correlations with prediction errors, as well as their discriminative power for ID and OOD samples are examined.
    \item[--] \name{}, a smoothness-based OOD detection strategy, ensures selective and reliable predictions on the ID inputs that are identified by their proximity to smooth domain regions.
    \item[--] \name{} is evaluated as a novel criterion for the systematic switch between the surrogate and its HF model counterpart in hybrid optimization settings, and aside from assessing the decrease in errors, overhead costs are also measured.
    \item[--] The assessment of FNN surrogates, local sensitivity profiles and \name{} approach, was conducted on three aircraft design variables study cases, along with comparisons to baselines, respectively, GPs, uncertainty estimates, and hybrid GP relying on deviations from design neighbors.
\end{itemize}
\textbf{The remainder of this paper is organised as follows.}\\ Section~\ref{sec:related_work} introduces the relevant concepts and literature in relation with our approach. Section~\ref{sec:approach} presents our novel smoothness-based OOD detection approach for NN-based surrogate models. Section~\ref{sec:evaluation} reports evaluation results, while Section~\ref{sec:conclusion} concludes the paper.

\section{Background and Related Works}
\label{sec:related_work}
This section provides background information about surrogate modeling for aircraft design and OOD detection strategies.
\subsection{Surrogate Modeling for Aircraft Design}
Surrogate models are data-driven and low-cost substitutes for the exact evaluation of the data points in design space, sometimes dominating the entire optimization process~\cite{ahmed2009surrogate}, and sometimes serving just as a supplementary aid to speed up the computations~\cite{sobester2008engineering, sobieszczanski1997multidisciplinary}.
Multiple techniques have been proposed in the literature to build data-fit surrogates that are trained with regression of high-fidelity simulation data. They usually rely on methods such as gaussian processes~\cite{simpson2001kriging,kuya2011multifidelity,kontogiannis2017multi}, proper orthogonal decomposition~\cite{iuliano2013proper}, eigenvalue decomposition~\cite{pagliuca2017model}, artificial neural network~\cite{linse1993identification, rai2000aerodynamic}, and more advanced techniques such as combining GP and neural networks~\cite{rajaram2020deep}. In the context of aircraft design optimization, the selection of the suitable surrogate modeling technique depends on the complexity of the underlying design problem, the ease of collecting high-fidelity simulation data, and the cost of development and maintenance. Surrogate modeling solutions are usually presented in conjunction with specific MDO problems from different industries. Hence, advanced techniques pose challenges in regards to the engineering efforts required to adapt them. This paper examines established data-driven modeling methods such as GP and FNN that can be applied directly to a wide variety of MDO problems, including aircraft design optimization. We developed a variety of surrogate models for different quantities of interest (QoI) using mainstream DL frameworks and GP modeling tools. These QoIs capture certain performance factors of an aircraft and are dependent on the operating flight conditions and the design variables. Our focus is on their deployments to accelerate the investigation of the design space and to find optimum solutions in hybrid surrogate optimization settings~\cite{zhao2011metamodeling, kipouros2007investigation, pagliuca2019surrogate}, where we are able to exploit information coming from both the original model and its surrogate. Therefore, reliable surrogate models with selective predictions (i.e., they are provided only under high-confidence conditions) are essential to ensure the high quality of these accelerated aircraft design optimizations.
\subsection{Out-of-Distribution Detection Strategies} 
A trustworthy surrogate model should not only produce accurate predictions based on known context, but also detect unknown inputs and forward them to the High-Fidelity model for safe handling. Indeed, Machine learning (ML) surrogates learn statistically from the available samples of data under the closed-world assumption~\cite{yang2021generalized}, where the test data is assumed to be drawn i.i.d. from the same distribution as the training samples, known as in-distribution (ID). However, when surrogate models are deployed in concrete aircraft design optimization, the sampled design configurations may represent inputs that are substantially different from the training samples, and should be considered as out-of-distribution (OOD). A wide range of detection strategies~\cite{bulusu2020anomalous, salehi2021unified} has been released to overcome the out-of-distribution challenge. Deep generative models~\cite{choi2018waic, nalisnick2018deep, pidhorskyi2018generative} have been leveraged to model efficiently the distribution of inputs $p(x)$ on the training samples. Then, a membership test is performed on any input $x$: if $p(x)$ is low, $x$ will be assigned to $\in\mathcal{D}_{ood}$, and vice versa. However, this generative modeling faces limitations on large-scale and/or complex input distributions. In contrast, model-dependent OOD detection methods focus on the distribution of model's intricacies, such as hidden features, softmax probability outputs, and uncertainty scores, rather than directly modeling input distributions~\cite{hendrycks2018deep}. Several model-dependent methods~\cite{yang2021generalized, salehi2021unified} have been proposed for classification neural networks. A few research works have focused on regression neural networks through thresholding on uncertainty estimates~\cite{kuleshov2018accurate}. Bayesian approaches~\cite{blundell2015weight, gal2017concrete, kingma2015variational} approximate the posterior distribution of neural networks parameters through an ensemble of models. Regarding Non-Bayesian ensembling approaches, Kendall and Gal~\cite{kendall2017uncertainties} adds the expected prediction's variance as additional network's output. Hence, the variance of a prediction is also improved by the learning algrithm, guided by the minimization of the negative log likelihood on the data. Mi et al.~\cite{mi2019training} proposed training-free uncertainty estimation that leverages the average of output deviations under input or feature map perturbations as a surrogate for uncertainty measurement. One challenge with these uncertainty-based methods is their reliability degeneration on out of distribution inputs on which they may be falsely overconfident~\cite{ovadia2019can}. Therefore, OOD examples can be identified by their predicted variances beyond the confidence interval obtained for ID samples. In this paper, we propose a novel OOD detection strategy for regression models under conditions of smoothness that relies on the mapping function's sensitivity profiling at a given input region to capture potentially OODs.   
\section{Approach}
\label{sec:approach}
\begin{figure*}[h]
\centering
\includegraphics[scale=0.7]{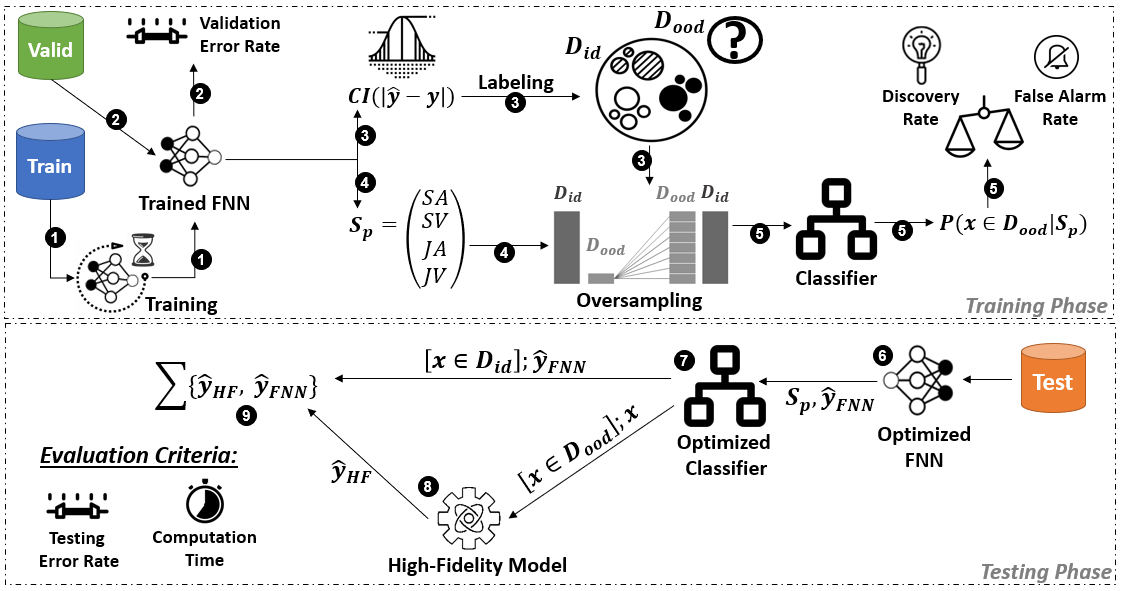}
\caption{Schema of \name{}: Co-Design of FNN Surrogate and Classification-based OOD Detection Models.}
\label{fig:SmOOD_Schema}
\end{figure*}

In this section, we describe the development steps required to codesign smoothness-based OOD detectors with FNN surrogates. In Figure~\ref{fig:SmOOD_Schema}, the proposed systematic workflow of the SmOOD approach is shown, including the OOD recognition, the training and testing of the inherent regression and classification models.
\subsection{Characterization of Out-of-Distribution}
\label{sec:Labeling_OOD}
In supervised learning, statistical models are commonly estimated via empirical risk minimization (ERM)~\cite{goodfellow2016deep}, a principle that considers minimizing the average loss on observed samples of data, as an empirical estimate of the true risk, i.e., the expected true loss for the entire input distribution. Unfortunately, ERM assumes that training and test data are identically and independently distributed (a.k.a. i.i.d. assumption). Distributional shifts often occur in real scenarios for many reasons, such as domain transition, temporal evolution, or selection bias, which degrades the model’s performance since certain captured correlations may not hold on these shifted inputs. Even worse, several research works~\cite{duchi2021learning, arjovsky2019invariant, liu2021heterogeneous} demonstrated that the optimized models can fail dramatically when involving strong distributional shifts. In many surrogate modeling cases where the HF simulations are expensive, selection bias is almost unavoidable, which ruins the i.i.d. assumption. The design of OOD detection methods for selective prediction is of more critical significance than reducing empirical risk further on the collected training samples. The first step consists of the OOD inputs characterization. 
In the context of aircraft surrogate performance models, we aim to characterize the foreseeable design configurations that can be sampled during the design optimization, however, they may deviate from the training distribution. As explained above, ERM optimizes the model by minimizing the average loss, which leads to greedily absorbing the correlations and patterns that hold on the majority of training instances. Hence, ERM can produce biased models that are susceptible to outliers, unfair to minor subsets of data, or prone to out-of-distribution samples~\cite{li2020tilted}. Thus, it is common to use held-out validation data as a proxy for unseen data points, aiming at more realistic estimates of the true risk. As shown in Figure~\ref{fig:SmOOD_Schema} (step $3$), the validation error rate may reveal the model's inefficiencies in terms of remaining high-error test inputs on which the learned patterns could not generalize, and they can be used for approximating the boundaries of the in-distribution. We rely on thresholding over the prediction confidence interval, i.e., the margin of validation errors at a certain level of confidence, and labeling the data point as OOD if its associated error exceeds the estimated margin.
\subsection{Computation of Local Sensitivity Profiles}
HF mathematical models are typically based on conservation laws and solve a coupled nonlinear system of partial differential equations on a discretized spatio-temporal domain. This allows numerically stable simulations of the aircraft via their approximations in smooth or piecewise smooth dynamics. Hence, their NN-based surrogate counterpart must meet this apriori of smoothness in regards to the predicted quantities of interest in order to be a viable alternative to the HF model. In fact, the smoothness of the FNN's mapping function affects the model complexity as evidenced by the fact that smooth deep neural networks tend to generalize better than their less smooth counterparts~\cite{novak2018sensitivity}. Nonetheless, rigid smoothness techniques that constrain the learned function excessively by forcing it to be equally smooth throughout the input space, may throw away useful information about the input  distribution~\cite{rosca2020case}. As alternatives to these rigid methods, DL practitioners can employ regularization techniques, such as weight decay~\cite{loshchilov2017decoupled}, dropout~\cite{srivastava2014dropout}, and early stopping~\cite{krizhevsky2012imagenet}, to encourage smoothness of the model and improve generalization. Their hyperparameters should be tuned to achieve the desired level of smoothness that allows the network to allocate capacity as needed to maintain useful diversity, handle input modalities, and capture task-relevant information. 
In our study of OOD detection for aircraft surrogate performance FNNs, the smoothness is one of the fundamental properties of system design. Apart from regularizing the FNNs to smooth mapping functions, we created pointwise sensitivity profiling to capture the local function smoothness around the neighbor regions of each evaluated data point. On unseen data points, the integrity of the surrogate network can then be ensured by comparing their triggered pointwise sensitivity profiles to the degree of smoothness observed on the optimized DL model on the in-distribution inputs. In the following, we introduce the proposed pointwise sensitivity profiling for FNN that quantify different aspects of the network's output variations in relation with changes in input variables. 
\subsubsection{The expected deviations of the network’s output:}
The straightforward way to estimate the sensitivity of the neural network at an individual data point, is to assess the induced output deviations in response to random noise injected into the input features. According to Mi et al.~\cite{mi2019training}, network sensitivity to input perturbations can be used as a surrogate for uncertainty. More precisely, they have proved that sensitivity and uncertainty have a nonnegative correlation in a setting of dense regression networks. Therefore, we include the following statistics on the output deviations under constrained input perturbations in the sensitivity profiling that will be used to separate between OOD and ID inputs.\\
Let $\Delta x$ be the perturbation applied to the input of $f$, a neural network, $\delta$ be the maximum threshold of the absolute value of $|\Delta x|$, and $\Delta y$ be the responding output deviation, $|f(x+\Delta x) - f(x)|$. Then, the two deviation-based sensitivity profiling metrics are defined respectively:
\begin{equation*}
\text{SA}(x) = \text{AVG}(|\Delta y|) = \mathbb{E}_{\Delta x\sim\mathcal{U}(-\delta,\delta)}[|f(x+\Delta x) - f(x)|]
\end{equation*}
\begin{equation*}
\text{SV}(x) = \text{VAR}(|\Delta y|) = \mathbb{E}_{\Delta x\sim\mathcal{U}(-\delta,\delta)}[(|\Delta y| - \text{AVG}(|\Delta y|))^2] 
\end{equation*}
\subsubsection{The input-output jacobian norm:}
As the feedforward neural networks are differentiable models, we also investigate its sensitivity at a point through the computation of input-output jacobian norm. Novak et al.~\cite{novak2018sensitivity} have presented extensive experimental evidence that the local geometry of the trained function as captured by the input-output Jacobian can be informative of the prediction confidence at the level of individual test points, and that it varies drastically depending on how close to the training data manifold the function is evaluated. Thus, the computed jacobian norm is likely to be higher at shifted inputs in comparison with the training inputs. Nevertheless, the feedforward networks are typically based on ReLU activations, which makes them not continuously differentiable. The derivative of the mapping function may therefore fluctuate sharply at small scales, and hence, considering the Jacobian of a specific data point will have less meaning than considering the Jacobians of a subset of nearby data points~\cite{smilkov2017smoothgrad}. Thus, we apply random perturbations on the data point to yield samples from its neighborhood. Then, we compute statistics on the vanilla Jacobians estimated at these noisy samples to be included in our network’s sensitivity profiling, as follows.\\ 
Let $\mathbf{J}(\mathbf{x})=\partial \mathbf{f}(\mathbf{x}) / \partial \mathbf{x}^{\mathbf{T}}$ be the input-output jacobian of a neural network, $f$, at the input $x$, and $||.||_F$ is the Frobenius norm. Further, we assume that $\Delta x$ is the perturbation applied to the input of $f$, a neural network, $\delta$ is the maximum threshold of the absolute value of $|\Delta x|$, and $x' = x + \Delta x$. 
\begin{equation*}
\text{JA}(x) = ||\text{AVG}(\mathbf{J}(\mathbf{x + \Delta x}))||_F = ||\mathbb{E}_{\Delta x\sim\mathcal{U}(-\delta,\delta)}[\mathbf{J}(x+\Delta x)||_F
\end{equation*}
\begin{equation*}
\text{JV}(x) = ||\text{VAR}(\mathbf{J}(\mathbf{x'}))||_F = ||\mathbb{E}_{\Delta x\sim\mathcal{U}(-\delta,\delta)}[(x' - \text{AVG}(\mathbf{J}(x')))^2||_F
\end{equation*}
As demonstrated in the step 4 of our defined workflow (Figure~\ref{fig:SmOOD_Schema}), we define the pointwise sensitivity profiles, denoted $S_p$, as a real-valued vector that concatenate the above senstivity-related statistics within the neighborhood regions of individual data points, $x$, which can be formulated as follows, $S_p(x) = [\text{SA}(x) \mathbin\Vert \text{SV}(x) \mathbin\Vert \text{JA}(x) \mathbin\Vert \text{JV}(x)]$.
\subsection{Training of ID/OOD Classifier}
After being estimated, the pointwise local sensitivity profiles are fed into a ML-based classifier, specifically, a binary classifier. The latter is trained on the sensitivity profiles of validation data points, along with their derived labels, i.e., ID or OOD, on the basis of the expected confidence interval for the validation errors. Nevertheless, the average and the margin of error are two metrics that will drive the development and the selection of surrogate models. Hence, it is reasonable to expect that these measurements are well optimized and only a small fraction of data points would remain with relatively high error over successive validation steps, which results in an imbalanced dataset of sensitivity profiles that is predominantly composed of ID examples with only a low percentage of OOD examples. To deal with class imbalance, we explore several commonly-used oversampling techniques~\cite{kovacs2019empirical} in the literature that produce synthetic data points belonging to the minority class to emulate a semblance of balance to the dataset. These techniques are necessary means of increasing the sensitivity of a classifier to the minority class, allowing the detection of as many OODs as possible from incoming inputs. This is shown in-between the step 4-5 of the SmOOD co-design workflow in Figure~\ref{fig:SmOOD_Schema}. By interpreting OODs as positive and IDs as negative, we should conduct an in-depth performance assessment of the ML-based classifier with aim of converging to optimal balance (i.e, the evaluation step 5 in Figure~\ref{fig:SmOOD_Schema}) between OOD discovery, referring to the ratio of the truly detected OODs among the actual OOD collection, and OOD false alarms, defined as the ratio of false positive arising from the actual ID samples. In the context of hybrid surrogate design optimization, a high OOD discovery rate ensures trustworthy selective surrogate predictions where only ID samples are submitted to the surrogate model, and hence, the accelerated design optimization converges to high potential and feasible aircraft design configurations. On the other hand, a low OOD false positive rate is imperative to preserve the advantages of surrogate modeling and avoid useless, costly queries to the HF model.
\subsection{\name{}: Integration and Evaluation}
During the training phase of the SmOOD workflow (Figure~\ref{fig:SmOOD_Schema}), the surrogate model is learned first, then its predictive errors are computed on the held-out validation datasets. Next, we leverage the confidence intervals of the estimated errors during the validation to identify the out-of-distribution (OOD) data (i.e., inputs on which the best-fitted model still produces relatively high errors). In parallel, we calculate the local sensitivity profiles for all the validation inputs. Since the best-fitted model was selected for its predictive performance, OOD inputs would have very small sensitivity profiles compared with their counterparts assigned to in-distribution (ID) inputs, and an oversampling procedure is required to amplify synthetically their occurrences in order to analyze the differences in smoothness-related characteristics between both groups of profiles. As a next step, we feed the oversampled sensitivity profiling data into a binary classifier that learns to distinguish between OOD and ID profiles. For optimal classifier selection, we assess both the OOD discovery and false alarm rates using k-fold cross-validation.\\
For the testing phase in the SmOOD workflow (Figure~\ref{fig:SmOOD_Schema}), the test inputs represent aircraft design configurations for which an assessment has been requested. Since SmOOD is a model-dependent OOD detection strategy, we first pass all the test inputs by the optimized FNN to determine their predictions and their corresponding local sensitivity profiles. Then, the optimized classifier serves as a calibrated OOD detector that generates a risk score for each precomputed sensitivity profile, quantifying the likelihood that its source test is indeed an out-of-distribution sample from the FNN surrogate's perspective. The next step depends on a criterion that evaluates whether the risk score falls below a predefined threshold (i.e., a default of  $0.5$ out of $[0, 1.0]$ score ranges), and two scenarios are then possible. If the criterion is satisfied, the surrogate prediction is returned and no further action is taken. If not, the test sample is sent to the High-fidelity model for simulation and its outcome is returned. Quality of results is assessed using two evaluation criteria. First criterion is the decrease rate achieved in respect to the prediction errors on testing samples, and especially, the further error reduction by the hybrid mode. A reliable OOD detection method must route all high risk samples to the HF model, reducing error rates significantly. Second, the computation time of hybrid mode should be much shorter than the pure HF-driven optimization since this was the main purpose of setting up the surrogate model in the first place. Indeed, a well-calibrated OOD detection method should spawn a low rate of false alarms because these false positives would cause unnecessary HF simulations, slowing down the progress towards design optimization.
\section{Evaluation}
\label{sec:evaluation}
In this section, we first introduce the three aircraft performance case studies, as well as our evaluation setup, metrics, and methodology. Next, we evaluate \name{} against standard baseline in terms of predictive performance and computation runtime.
\subsection{Experimental Setup}
This section details the different elements that were set up for SmOOD performance assessment, including aircraft design study cases, inherent models, baseline, as well as evaluation environment, strategy, and metrics.
\subsubsection{Case Studies}
Below, we briefly describe the studied aircraft performance factors~\cite{anderson1999aircraft}, the influencer design variables~\cite{anderson1999aircraft}, and the design of experiments used for dataset collection.\\
\textbf{Maximum TakeOff Weight (MTOW).} It represents the maximum weight at which the pilot of the aircraft is allowed to attempt to take off given its structural design. MTOW is an important factor in aircraft design. A higher MTOW means the aircraft can take off with more fuel and has a longer range, which makes it more appealing to customers. However, MTOW is subject to several structural constraints during the design optimization process.\\
\textbf{Time To Climb (TTC).} Climbing is the act of increasing the altitude of an aircraft. The main climb phase is the increase of altitude following the takeoff to reach the cruise level. As a way to measure an aircraft’s climb performance, it is common to set up a reference cruise altitude level, then, estimate the time needed to climb to the predetermined altitude at a constant airspeed. This climb performance measurement is called the Time To Climb (TTC).\\
\textbf{Balanced Field Length (BFL).} It refers to the shortest runway length at which a balanced field takeoff can be performed by an aircraft design while complying with safety regulations. A balanced field takeoff occurs when the required accelerate-stop distance is equal to the required takeoff distance. Accelerate-stop distance is the runway length required by an aircraft to accelerate to a specific speed, and then, in the event an engine fails, to stop safely on the runway. Thus, aircraft designers are senstitive to BFL as any changes in this require ensuring safety margins at takeoff are still respected.\\ 
\textbf{Design Configuration.} A set of 15 design variables are considered in the surrogate modeling based on experts’ judgment on their influence on the above-mentioned quantities of interest. The design variables capture the wing design (i.e., aspect ratio, taper ratio, thickness ratio, and winglet span ratio), the fuselage geometrical structure, the engine thrust, and the mass of aircraft parts.\\ 
\textbf{Design of Experiment(DoE).} Our engineering collaborators configure the appropriate design of experiments (DOE) to collect the labeled data points. DoE assembles a set of tests that exercises the HF model across diverse design configurations to gather HF simulations. Indeed, the Latin Hypercube Sampling technique~\cite{loh1996latin} is leveraged to efficiently sample from large, multivariable design spaces. LHS uniformly divides the range of each design variable into the same number of levels. It then systematically combines independent samples of the levels of each factor to specify a variety of random data points in the design space. Running the HF model on these data points gives us HF simulations that map design factors to their corresponding responses, i.e., the measurable quantities of interest. Therefore, we obtain $2259$ of training data points and $1960$ of testing data points for each one of the study cases.
\subsubsection{Models}  
\hfill \\
\textbf{Surrogate.} Our base nonlinear regression model is a three-layer feedforward neural network(FNN)~\cite{bebis1994feed} that is trained using the Mean Squared Error (MSE) loss function with L2-norm regularization. Rectified linear units (ReLU) are used as hidden layer activation functions, and Adam is used as the optimization algorithm. In regards to the architecture, we followed the design principle of pyramidal neural structure~\cite{design_principles}, i.e., from low-dimensional to high-dimensional feature spaces/layers, as well as these dimensions are powers of $2$ to achieve better performance on GPUs~\cite{nvidia}. Regarding the hyperparameters tuning, we leverage the grid-search strategy and 5-fold cross validation to sample and try several possible settings. To outline their ranges, we denote $linspace(a, b, n)$ to indicate the range of $n$ equi-spaced values within $[a, b]$ and $logspace(c, d, base)$ to indicate the interval of ${base^c,.., base^d}$, where $c < d$. The FNN’s width of layers are selected from $logspace(5, 10, 2)$, then, the optimizer's hyperparameters were tuned as follows: learning rate $\eta \in s \cup 3\times s$, weight decay $\lambda \in s \cup 5\times s$, where $s = logspace(-4, -1, 10)$. Batch size was tuned in $logspace(3, 7, 2)$, and epochs count in $linspace(50, 500, 50)$.\\    
\textbf{OOD Detection.} Several ML classifiers (Logistic Regression~\cite{wright1995logistic}, SVM~\cite{noble2006support}, Random Forest~\cite{biau2016random}, and Gradient Boosting~\cite{natekin2013gradient}) and over-sampling techniques (SMOTE~\cite{chawla2002smote}, BorderlineSMOTE~\cite{han2005borderline}, ADASYN~\cite{he2008adasyn}, and SVMSMOTE~\cite{tang2008svms}) have been evaluated on our OOD detection problems using 5-fold cross validation process. We have found that Gradient Boosting and SVMSMOTE are the best design choices. Then, we follow the same hyperparameters tuning method used for FNN, i.e.,  grid-search strategy and 5-folds cross validation. For the gradient boosting classifier, we examine the learning rate, $\eta$ and the number of estimators,$n$, selected from the following ranges, respectively, $s\cup 5\times s$ and $[100, 250, 500]$, where $s = logspace(-3, -1, 10)$ and $[100, 250, 500]$. In regards to SVMSMOTE oversampler, we consider $k\in[5, 10,15]$, where $k$ represents the count of nearest neighbors to used to construct synthetic samples, and $r\in[0.25, 0.5, 0.75, 1.0]$, where $r$ indicates the desired ratio of the number of samples in the minority class over the number of samples in the majority class after resampling. 
\subsubsection{Baseline}
\label{sec:baseline}
\hfill \\
\textbf{Surrogate.} We compare our FNNs with Gaussian Process(GP) regression that is already well researched for surrogate modeling~\cite{wendl2004scattered}, replacing expensive high-fidelity aircraft simulations. A GP~\cite{krige1951statistical} is a generalization of the Gaussian distribution to describe universal functions $f(x)$. The main ingredient of GP design is the selection of the covariance function $k(x, x')$, also called kernel. For our comparison with FNN, we are interested in approximating multiple quantities using the baseline. We chose the radial basis function (RBF)~\cite{buhmann2003radial} that has been implemented to interpolate several aircraft design data~\cite{kontogiannis2017multi, pagliuca2019surrogate}. RBF~\cite{buhmann2003radial} is an universal kernel function that can be used to fit any complex non-linear regression data. 
The hyperparameters of the kernel are optimized during fitting by maximizing the log-marginal-likelihood (LML) on the training data. As the LML may have multiple local optima, we set up $10$ to be the number of repetitive restarts of the optimizer in order to improve the convergence towards optimal results.\\
\textbf{OOD Detection.} A basic approach of hybrid GP~\cite{pagliuca2019surrogate} for surrogate aircraft design optimization mimics a rule of thumb in engineering analysis: if the new data is reasonably similar to the existing one, it can be assumed to be reliable. Given a design configuration to assess, one can compute the standard deviation, $\sigma$, of the differences between the output predicted by RBF, $\hat{y}$, and the actual outputs, $y_1, y_2,...,y_n$, of a set of $n$ neighboring configurations. The standard deviation $\sigma$ can be computed with differences $\left\|\mathbf{\hat{y}}-\mathbf{y}_{i}\right\|$ with $i\in[1,n]$, as follows,$\sigma=\operatorname{std}\left(\left[\left\|\mathbf{\hat{y}}-\mathbf{y}_{1}\right\|,\left\|\mathbf{\hat{y}}-\mathbf{y}_{2}\right\|,\left\|\mathbf{\hat{y}}-\mathbf{y}_{n}\right\| \ldots\right]\right)$. Then, the obtained standard deviation is compared to a threshold, $\sigma_t$. If it is lower, $\sigma<\sigma_t$, the prediction of RBF surrogate is adopted; otherwise, the HF model must be requested instead. Regarding the threshold value $\sigma_t$, a priori statistical analysis can be performed on the validation data to derive the threshold value that includes $95\%$ of configurations. Using a 5-fold cross validation, we tune the number of neighbors considered in the computation of standard deviation within the range of $linspace(4, 20, 4)$, and we select the number that yields the highest correlation between the estimated standard deviation and the actual error.
\subsubsection{Evaluation Procedure}
To evaluate SmOOD's effectiveness, we conducted quantitative and qualitative analyses. In the following, we detail the metrics and the procedure used.\\
\textbf{Quantitative. }Due to the high cost of running HF simulations, we adopt a $10$-fold cross-validation method for all experiments in order to have different splits for the training and validation datasets and quantify the target metrics by averaging their values over the $10$ iterations. Besides, all the included estimated metrics like error rates and runtime, are computed as average values over $5$ runs or more, in order to mitigate the effects of randomness inherent in statisical learnign algorithms. Below, we introduce the different evaluation metrics that have been used in the empirical evaluations.\\
\textit{NRMSE.} stands for normalized root mean square error, and it is scale-independent version of RMSE that allows comparison between models at different scales. Indeed, RMSE is the average deviation between predictions and actual outputs, measured on the same scale and with the same output unit, and can be formulated as follows: $RMSE=\sqrt{\sum_{i=1}^{n}\left(y_{i}-\hat{y}\right)^{2}/n}$, where $y_{i}$  is the ith observation of y and  $\hat{y}_{i}$ its corresponding prediction by  the model.\\
In NRMSE, the expected model deviations are reported relative to the overall range of output values, and the formula becomes $NRMSE = RMSE/(y_{\max }-y_{\min })$.\\
\textit{Confidence Interval (CI).} indicates the degree to which an estimate is expected to vary from the average, within a certain level of confidence. For example, a 95\%CI of a model's error represents the upper and lower bounds within which the estimate of error will fall for 95 percent of the time (i.e., 95 samples out of 100 were taken). We use the bootstrapping method to produce accurate CIs because its non-parametric nature allows it to be used without making any prior assumptions about the estimate distribution.\\
\textit{Precision.} represents the proportion of the positive identifications that were actually correct in a binary ML classification problem.\\
\textit{Recall.} measures the proportion of actual positives that were correctly classified in a binary ML classification problem.\\
\textit{Macro-Averages of Precision/Recall.} refer to the arithmetic averages of the precision and recall scores of individual classes.\\ 
\textit{Precision-Recall (PR) curve.} is a useful evaluation plot for binary classifiers that return class membership probabilities, when there is moderate to large class imbalance~\cite{davis2006relationship}. It simply plots the precision versus recall obtained by a classifier using different probability thresholds.\\
\textit{Area Under Precision-Recall (AUPR).} approximates the area under the PR curve, ranging from $0.0$ to $1.0$.\\
\textit{\%Decr\_Err.} The decrease ratio of error we achieve by a certain improvement, can be formulated as follows:\\
\begin{equation}
\label{decr_err}
\%Improv\_Err = \dfrac{\textit{pre-NRMSE} - \textit{post-Err}}{\textit{pre-NRMSE}} \times 100
\end{equation}
Where \textit{pre-NRMSE} and \textit{post-NRMSE} refer to the actual NRMSE and the reduced NRMSE.\\
\textit{Speed up.} is a popular measure for the relative performance of two systems processing the same problem. In our case, we denote $T_s$ be the surrogate compute time, and $T_o$ the High-fidelity compute time. Then, the speedup due to surrogate modeling can be computed as follows, $S_{pure} = T_o/T_s$. Given our OOD detection strategy used in hybrid surrogate optimization, we obtain $p$, as a proportion of instances that can be predicted by the surrogate. Then, the remaining $1-p$ proportion of instances that require requests to HF model. This means that the speed up of the hybrid surrogate model can be formulated as follows, $S_{hybrid} = T_o/(T_d + p\times T_s+(1-p)*T_o)$, where $T_d$ is the total computation time needed to predict whether each of the inputs is OOD or not.\\
\textbf{Qualitative. }In order to obtain the input of domain experts, we interviewed two senior aircraft engineers with more than 10 years of experience, who work with our industrial partner, Bombardier Aerospace. Both of them are proficient in the use of MDO in aircraft design. Furthermore, they are part of the team that developed the high-fidelity physics model; therefore, they are familiar with the QoIs associated with the case studies.\\
As a first step, we present the quantitative analysis to both aircraft engineers separately so that they can get detailed performance metrics, as well as all execution costs for all model development steps (training, tuning of hyperparameters, etc). Then, we ask them to provide a critique of the significance of the obtained performance metrics from an aircraft design standpoint. Specifically, we seek comparisons of FNN surrogates along with our co-designed OOD detection method, as opposed to conventional surrogate modeling setups with GP and uncertainty quantification. In addition, we request explanations on how SmOOD contributes to quality assurance and acceleration of data-driven aircraft design optimization. Afterwards, we compile and merge their opinions and comments into paragraphs. Next, we submit them for approval to ensure their claims are reflected in the written content, and that they both agree with the conclusions made. Final consensus statements are added as feedback from domain experts to each related research question.
\subsubsection{Environment}
We use Pytorch~\cite{paszke2019pytorch}, an established DL framework for modeling and training feedforward neural networks. We leverage GPy~\cite{gpy2014}, a popular and maintained framework to design and train Gaussian processes. In terms of the hardware environment, we use CPU machines for all the experiments for fairness comparisons. The HF models were running on a machine with Intel Xeon CPU E5-1630 v3 of 3.5Ghz and a 32 Gb of RAM. The FNNs and GPs were running on Standard Virtual Machines using 4 cores on Intel Xeon Platinum 8168 CPU of 2.70GHz and 8 Gb of RAM.
\subsection{Experimental Resuls}
In conducting the evaluation of the proposed approach, we studied the following research questions:
\subsubsection*{RQ1. Is FNN a viable general-purpose approximator to model complex surrogate aircraft design performance models?}
\hfill

\textbf{Motivation.} The objective is to evaluate the effectiveness of FNN as a universal approximator for surrogate aircraft design performance models in comparison to GP, which is the mainstream universal estimator for surrogate modeling. 
\begin{table*}[ht]
\centering
\caption{Performance Metrics for different pairs of QoI and Surrogate Model.}
\label{tab:perfs_comparison}
\begin{tabular}{|c|c|c|c|c|c|c|c|}
\hline
\multicolumn{1}{|l|}{QoI}   & \multicolumn{1}{l|}{Model} & \multicolumn{1}{l|}{Valid\_Err} & \multicolumn{1}{l|}{Test\_Err} & \multicolumn{1}{l|}{Valid\_CI\_Err@99\%CL} & \multicolumn{1}{l|}{Test\_CI\_Err@99\%CL} & \multicolumn{1}{l|}{\%Valid\_OODs} & \multicolumn{1}{l|}{\%Test\_OODs} \\ \hline
\multirow{2}{*}{\textbf{MTOW}} & \textit{GP}                & 0.0415                          & 0.0363                         & {[}0.0212,0.0596{]}                        & {[}0.014, 0.0547{]}                       & 5.14\%                             & 4.39\%                            \\ \cline{2-8} 
                               & \textit{FNN}               & 0.0353                          & 0.0319                         & {[}0.0163, 0.0523{]}                       & {[}0.0111, 0.0486{]}                      & 1.59\%                             & 1.17\%                            \\ \hline
\multirow{2}{*}{\textbf{TTC}}  & \textit{GP}                & 0.1707                          & 0.1698                         & {[}0.1646, 0.1769{]}                       & {[}0.1632, 0.1762{]}                      & 40.95\%                            & 41.33\%                           \\ \cline{2-8} 
                               & \textit{FNN}               & 0.0764                          & 0.0803                         & {[}0.0662, 0.0865{]}                       & {[}0.0668, 0.0942{]}                      & 5.0\%                              & 5.26\%                            \\ \hline
\multirow{2}{*}{\textbf{BFL}}  & \textit{GP}                & 0.0454                          & 0.0439                         & {[}0.0241, 0.0637{]}                       & {[}0.0168, 0.0664{]}                      & 6.64\%                             & 5.97\%                            \\ \cline{2-8} 
                               & \textit{FNN}               & 0.0435                          & 0.0361                         & {[}0.0198, 0.0637{]}                       & {[}0.0124, 0.0599{]}                      & 1.42\%                             & 1.12\%                            \\ \hline
\end{tabular}
\end{table*}
\textbf{Method.} We train and tune both FNN and GP surrogate models on all the three aircraft design performance factors. Then, we test the optimized models on a held-outs test dataset to assess their predictive performance. As performance metrics, we compute the NRMSE on validation datasets to gauge the fitness quality of both models and the NRMSE on test dataset to compare their generalizability capabilities. Besides, we infer the CI of each estimate of error with 99\% of confidence level, as an accurate risk assessment of the possible range of the model’s prediction errors, and especially, the upper bound that indicates how severe the error is expected to be. Last, we calculate the ratio of OOD inputs observed on validation and testing datasets according to our OOD labeling rule (Section~\ref{sec:Labeling_OOD}), i.e., their actual prediction errors are higher than the maximum expected margin.

\textbf{Results.} Table~\ref{tab:perfs_comparison} summarizes the obtained performance metrics for each pair of regression problem and surrogate model type. The results show that FNNs have reached lower validation and test estimation errors with even tighter confidence intervals than GP in all the studied surrogate modeling problems. Moreover, the ratio of model-dependent OOD inputs that have been experienced by GP is substantially higher than FNN. This demonstrates that GP's high average error and wide confidence interval are due to the prevalence of complex design configuration inputs on which GP fails and cannot provide a reliable assessment. It is on the time-to-climb (TTC) design problem that the gap between the two types of surrogate models is most pronounced. There are, indeed, high nonlinearities in TTC modeling, where two numerically-close inputs can result in distant takeoff behaviors. Because GP kernels are biased towards their predefined distributional priors and have a limited learning capacity in comparison to FNN, they do not generalize well to aircraft performance design problems with dispersed behaviors, such as the time to climb simulations, ranging from takeoff failures to accelerated climbing scenarios.

\FrameSep.5em
\begin{framed}
\noindent
\textbf{Finding 1:} Feedforward neural network is a viable universal approximator, outperforming Gaussian Process, in building complex, highly-nonlinear surrogate aircraft performance models.
\end{framed}

\textbf{Domain Expert Feedback.} Aircraft engineers find the comparison results enlightening as they demonstrate that data-driven surrogate modeling is still a viable solution even in cases of high nonlinearity input-output mappings, such as TTC models. Moreover, GP's results on TTC are expected, since it is the origin of this initiative on reliable FNN surrogates. It was previously necessary for aircraft engineers to re-run the design optimization with the HF model in the objective function because the GP gives inaccurate TTC predictions and hinders the convergence of the optimizer, which does not find a feasible design.

\subsubsection*{RQ2. Can local sensitivity profiles capture relevant information on FNN behaviors to accurately signal OODs across generated samples?}
\hfill

\textbf{Motivation.} The aim is to determine how well pointwise local sensitivity profiles can discriminate between ID and OOD inputs compared with exploiting the uncertainty estimation of GP.

\textbf{Method.} For GP, the predictions, $\hat{y}$ are made in a probabilistic way, i.e., described by mean, $\mu_{\hat{y}}$, and standard deviation, $\sigma_{\hat{y}}$, which is supposed to reflect the epistemic uncertainty, i.e., how certain the model is with respect to its prediction. We therefore statistically compare the distribution of $\sigma_{\hat{y}}$ on ID versus OOD examples discovered during the validation. To do that, we compare their box plots and perform Mann-Whitney U nonparametric hypothesis tests to determine whether the $\sigma_{\hat{y}}(\mathcal{D}_{id})$ and $\sigma_{\hat{y}}(\mathcal{D}_{ood})$ were sampled from the same population. Regarding FNN, we compute local sensitivity profiles triggered by FNN on the validation data points, and then analyze the distribution of these profiles among the ID and OOD groups. To do that, we use Principal Components Analysis (PCA) to visualize the computed multivariate local sensitivity profiles into 2D graphs. Indeed, we reduce the dimensions into two orthogonal principal components that preserve the maximum amount of variance explained by the original multivariate points, as well as, we add the original variable vectors to show their correlations and directions w.r.t the principal components. Moreover, we colored the ID and OOD points with different colors to identify the differences between the two groups. In addition, we perform Mann-Whitney U tests for each of the sensitivity profile variables to statistically confirm if the ID and OOD groups are likely to yield values originating from two different distributions.
\begin{figure*}[h]
\centering
\includegraphics[scale=0.6]{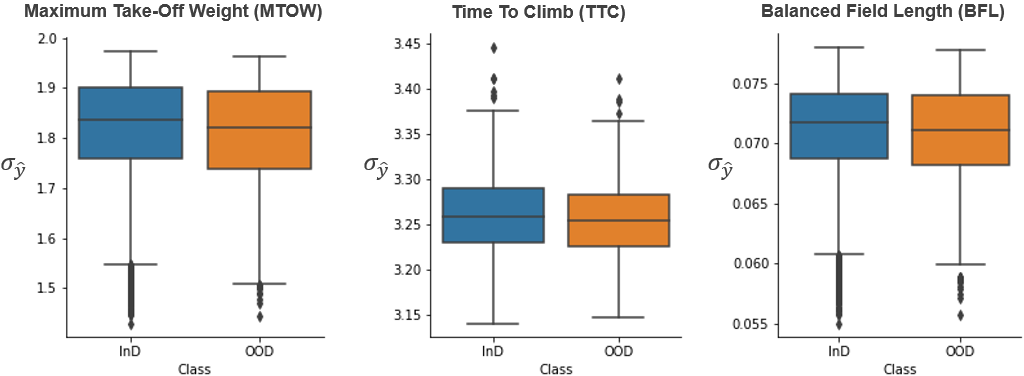}
\caption{Comparison of GP's standard deviation distribution between ID and OOD samples for each QoI}
\label{fig:stds_comp}
\end{figure*}
\begin{figure*}[h]
\centering
\includegraphics[scale=0.7]{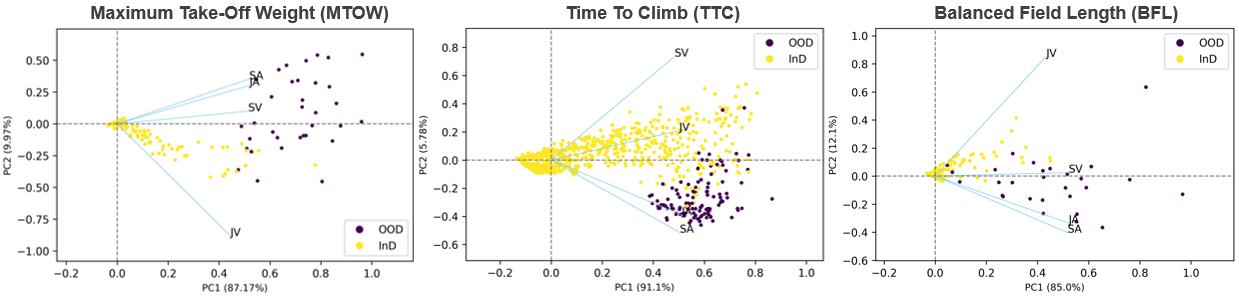}
\caption{Graph of Pointwise Sensitivity Profiles for $\mathcal{D}_{id}$ and $\mathcal{D}_{ood}$ w.r.t the derived 2 Principal Components}
\label{fig:SPs_comp}
\end{figure*}

\textbf{Results.} Figure~\ref{fig:stds_comp} shows the box plots of the GPs’ standard deviations for both ID and OOD examples. As can be seen, the groups' boxes have very close sizes and almost overlap completely for all the regression problems. Additionally, hypothesis testing suggests no statistically significant differences between the uncertainty estimates predicted by GP on ID versus OOD. The median lines in the box plots indicate that the median of the standard out-of-distribution sample is lower than the standard identification sample, demonstrating how GP incorrectly produces a sense of overconfidence about out-of-distribution samples. Even worse, the comparison of the median lines within the boxes highlights that the median of $\sigma_{\hat{y}}(\mathcal{D}_{ood})$ is lower than the median of $\sigma_{\hat{y}}(\mathcal{D}_{id})$, which shows how GP incorrectly tends to produce overconfident epistemic uncertainty when confronted with out-of-distribution samples. This is definitely an expected degradation in the reliability of uncertainty estimates, since the posterior distributions of predictions were calibrated using available samples, mostly ID, suggesting bias against distant examples. In contrast, Figure~\ref{fig:SPs_comp} present 2D plots of PCA analyses based on the sensitivity profiles of the regression models for the studied QoIs, MTOW, TTC, and BFL. As can be observed, the ID examples are grouped together and form a sort of cluster, whereas the OOD examples are quite dispersed and far from the centroid of the ID group. The hypothesis testing results confirm these observations, and there are statistically significant differences (with \textit{p-value} $< 1e-6$) between the ID and OOD data groups in regards to all of the sensitivity profile's variables. Furthermore, the high dispersion and rarity of the OOD data points highlight that the identified scenarios represent corner-case and extreme behaviors with diffent patterns, rather than representing a novel distribution. 

\FrameSep.5em
\begin{framed}
\noindent
\textbf{Finding 2:} Pointwise local sensitivity profiles can accurately separate OOD and ID samples using FNN, whereas the reliability of GP’s epistemic uncertainty declines on OOD samples.
\end{framed}

\textbf{Domain Expert Feedback.} Aircraft engineering experts confirm they have experienced overconfidence of GP uncertainties before, especially when using objective functions in MDO that include the variance to better steer the search to high potential regions. In that case, over-confident variance is misleading to the optimizer. They perceive the added value of the FNN sensitivity profiling, which is able to capitalize on the smoothness of input-output mapping as it is an a priori system property and an implicit assumption in the HF model’s differential equations.
\subsubsection*{RQ3. How effective is \name{} in the coverage of OOD inputs across generated samples?}
\hfill

\textbf{Motivation.} The goal is to assess the performance of \name{} when co-designed with FNN against the baseline strategy used with GP in terms of precision and coverage.

\textbf{Method.} We train and tune our gradient boosting classifier on the local sensitivity profiles yielded by FNNs during the validation for each regression problem. Then, we test the optimized classifier on the held-out testing dataset. Next, we draw PR curves along with their associated AUPR scores for each studied problem. Concerning the OOD detection baseline for GP introduced in Section~\ref{sec:baseline}, 
it provides only labels with no class scores, we compute the precision and recall scores for each class, separately, and their macro average for each pair of surrogate model and OOD detector.
\begin{table}[ht]
\centering
\caption{Performance Comparison of SmOOD and Base for OOD Detection.}
\label{tab:ood_comparison}
\resizebox{180pt}{!}{%
\begin{tabular}{|c|c|c|c|c|c|}
\hline
Target                         & Detector                        & Model                         & Class & Precision & Recall \\ \hline
\multirow{6}{*}{\textbf{MTOW}} & \multirow{3}{*}{\textit{Base}}  & \multirow{3}{*}{\textit{GP}}  & ID    & 96\%      & 96\%   \\ \cline{4-6} 
                               &                                 &                               & OOD   & \textbf{11\%}      & \textbf{10\%}   \\ \cline{4-6} 
                               &                                 &                               & MA    & 53\%      & 53\%   \\ \cline{2-6} 
                               & \multirow{3}{*}{\textit{SmOOD}} & \multirow{3}{*}{\textit{FNN}} & ID    & 100\%     & 100\%  \\ \cline{4-6} 
                               &                                 &                               & OOD   & \textbf{72\%}      & \textbf{91\%}   \\ \cline{4-6} 
                               &                                 &                               & MA    & 86\%      & 95\%   \\ \hline
\multirow{6}{*}{\textbf{TTC}}  & \multirow{3}{*}{\textit{Base}}  & \multirow{3}{*}{\textit{GP}}  & ID    & 57\%      & 91\%   \\ \cline{4-6} 
                               &                                 &                               & OOD   & \textbf{14\%}      & \textbf{2\%}    \\ \cline{4-6} 
                               &                                 &                               & MA    & 36\%      & 47\%   \\ \cline{2-6} 
                               & \multirow{3}{*}{\textit{SmOOD}} & \multirow{3}{*}{\textit{FNN}} & ID    & 100\%     & 98\%   \\ \cline{4-6} 
                               &                                 &                               & OOD   & \textbf{70\%}      & \textbf{93\%}   \\ \cline{4-6} 
                               &                                 &                               & MA    & 85\%      & 95\%   \\ \hline
\multirow{6}{*}{\textbf{BFL}}  & \multirow{3}{*}{\textit{Base}}  & \multirow{3}{*}{\textit{GP}}  & ID    & 94\%      & 96\%   \\ \cline{4-6} 
                               &                                 &                               & OOD   & \textbf{13\%}      & \textbf{10\%}   \\ \cline{4-6} 
                               &                                 &                               & MA    & 54\%      & 53\%   \\ \cline{2-6} 
                               & \multirow{3}{*}{\textit{SmOOD}} & \multirow{3}{*}{\textit{FNN}} & ID    & 100\%     & 100\%  \\ \cline{4-6} 
                               &                                 &                               & OOD   & \textbf{76\%}      & \textbf{73\%}   \\ \cline{4-6} 
                               &                                 &                               & MA    & 88\%      & 86\%   \\ \hline
\end{tabular}}
\end{table}
\begin{figure*}[h]
\centering
\includegraphics[scale=0.6]{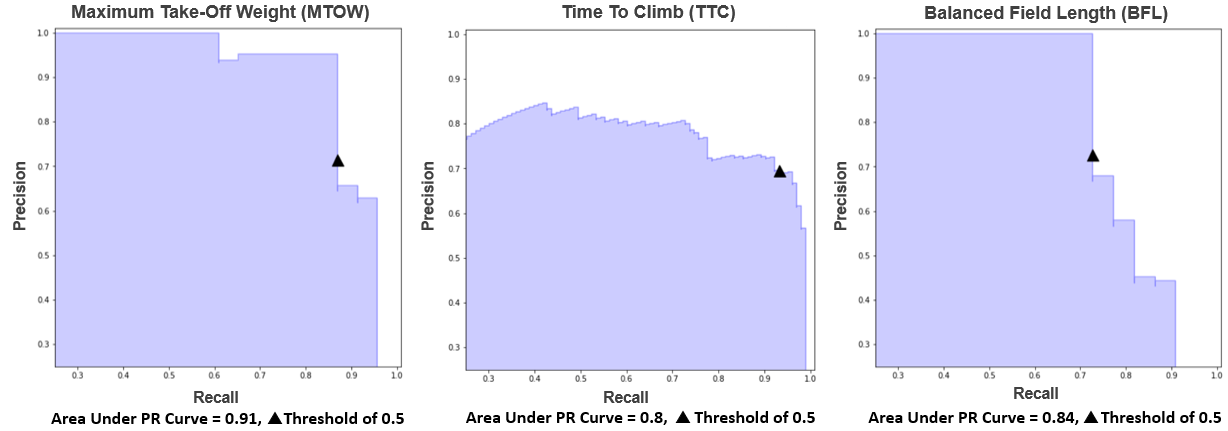}
\caption{Precision-Recall Curves obtained by the optimized binary classifier for each QoI}
\label{fig:pr_curves}
\end{figure*}

\textbf{Results.} Figure~\ref{fig:pr_curves} shows the precision-recall curves obtained by FNN for each aircraft performance problem and their associated AUPR scores. As can be seen, \name{} was effective in fitting the collected local sensitivity profiles as evidenced by the relatively high AUPR scores. Although some levels of precision could be sacrificed to increase recall, we believe it is better to leave the default threshold at $0.5$ since it yields a good trade-off between precision and recall, as shown by the marked points on the PR curves. Further pre-tuning of thresholds on validation dataset can lead to overfit the validation dataset and inversely causes a degradation of the classifier’s performance. Besides, Table~\ref{tab:ood_comparison} summarizes all the classification scores: both of precision and recall for each class, as well as their macro-averages for each pair of surrogate model type and aircraft design performance problem. Our method outperforms by far the baseline approach. This demonstrates the lack of generalizability expressed by the rigid smoothness apriori of the baseline OOD detection method, which could not hold for complex, highly-nonlinear aircraft design performance models. Our method, in contrast, retrieves pointwise local sensitivity profiles and fits a binary ML classifier to the high-fidelity simulations to serve as a calibrator and detects erroneous smoothness behavior triggered by the FNN during the inference on unseen test inputs. 

\FrameSep.5em
\begin{framed}
\noindent
\textbf{Finding 3:} \name{} was effectively able to reveal at least $73\%$ and up to $93\%$ of the out-of-distribution examples on the different aircraft design performance models.
\end{framed}

\textbf{Domain Expert Feedback.} Aircraft engineers emphasize the importance of developing such mitigation strategies against behavioral drifts of these black-box ML models to prevent their use under suspicious conditions. They point out that design optimization studies are actually driven by an end-to-end automated process, which results in one best configuration communicated to managers and related engineering teams. Thus, such 3-5\% of unreliable predictions can enable the accelerated MDOs to reach feasible solutions at an early stage, however, the inefficiency of these solutions might remain hidden for quite a while.

\subsubsection*{RQ4. What are the benefits of deploying \name{} in hybrid surrogate optimization settings?}
\hfill

\textbf{Motivation.} We aim to examine the effects of \name{} on the prediction errors and computation time when it is combined with the FNN for hybrid surrogate aircraft design optimization.

\textbf{Method.} We compute the NRMSE and inference runtime values for all the surrogate variants: GP, FNN, HybridGP (GP + Baseline) and HybridFNN (FNN + SmOOD). Then, we derive their SpeedUp relative to the HF model, as well as, the Decrease rate of NRMSE achieved by the hybrid versions w.r.t. the pure surrogate counterparts. As a way to compare the surrogate model types, we calculated, separately, the computation durations required for all the design and evaluation steps (training, tuning, validation, and testing) for each type of model. Furthermore, we compute the same computation times for the design of OOD detection strategies to determine the overhead added when using an OOD detector to switch between HF model and its surrogate counterpart in hybrid surrogate design optimization settings.
\begin{table}[ht]
\centering
\caption{Improvement Evaluations for different pairs of QoI and Surrogate Model.}
\label{tab:improv_evaluations}
\resizebox{\columnwidth}{!}{%
\begin{tabular}{|c|c|c|c|c|c|}
\hline
QoI                   & Model     & Test\_ERR & \%ERR\_Decr & Runtime     & SpeedUp    \\ \hline
\multirow{5}{*}{\textbf{MTOW}} & HF        & -         & -           & 02h:57m:58s & 1          \\ \cline{2-6}   
                      & GP        & 0.0363    &    -         & 299.26ms    & $3.57\times10^4$ \\ \cline{2-6} 
                      & HybridGP  & 0.036     & \textbf{0.83\%}      & 07m:38s     & \textbf{$2.33\times10$}  \\ \cline{2-6} 
                      & FNN       & 0.0319    &   -          & 0.81ms      & $1.32\times10^7$ \\ \cline{2-6} 
                      & HybridFNN & 0.0169    & \textbf{47.02\%}     & 02m:38s     & \textbf{$6.76\times10$}  \\ \hline
\multirow{5}{*}{\textbf{TTC}}  & HF        & -         & -           & 02h:57m:58s & 1          \\ \cline{2-6} 
                      & GP        & 0.1698    &    -         & 289.47ms    & $3.69\times10^4$ \\ \cline{2-6} 
                      & HybridGP  & 0.1677    & \textbf{1.24\%}      & 10m:48s     & \textbf{$1.65\times10$}  \\ \cline{2-6} 
                      & FNN       & 0.0803    &  -           & 15.51ms     & $6.88\times10^5$ \\ \cline{2-6} 
                      & HybridFNN & 0.0441    & \textbf{45.08\%}     & 12m:31s     & \textbf{$1.42\times10$}  \\ \hline
\multirow{5}{*}{\textbf{BFL}}  & HF        & -         & -           & 02h:57m:58s & 1          \\ \cline{2-6} 
                      & GP        & 0.0439    &  -           & 290.07ms    & $3.68\times10^4$ \\ \cline{2-6} 
                      & HybridGP  & 0.0435    & \textbf{0.91\%}      & 08m:32s     & \textbf{$2.08\times10$}  \\ \cline{2-6} 
                      & FNN       & 0.0361    &   -          & 4.58ms      & $2.33\times10^6$ \\ \cline{2-6} 
                      & HybridFNN & 0.0317    & \textbf{12.19\%}     & 01m:54s     &  \textbf{$1.42\times10$} \\ \hline
\end{tabular}
}
\end{table}
\begin{table}[ht]
\centering
\caption{Computation Times for different steps of Surrogate Model Design.}
\label{tab:surrogate_step_times}
\resizebox{\columnwidth}{!}{%
\begin{tabular}{|c|c|c|c|c|c|}
\hline
QoI                            & Model & Train\_Time & Tune\_Time  & Eval\_Time & Test\_Time \\ \hline
\multirow{2}{*}{\textbf{MTOW}} & GP    & 02m:50s     & -           & 27m:28s    & 299.26ms   \\ \cline{2-6} 
                               & FNN   & 00m:17s     & 05h:41m:39s & 02m:53s    & 0.81ms     \\ \hline
\multirow{2}{*}{\textbf{TTC}}  & GP    & 04m:25s     & -           & 42m:07s    & 289.47ms   \\ \cline{2-6} 
                               & FNN   & 02m:10s     & 22h:49m:54s & 21m:04s    & 15.51ms    \\ \hline
\multirow{2}{*}{\textbf{BFL}}  & GP    & 03m:11s     & -           & 31m:12s    & 290.07ms   \\ \cline{2-6} 
                               &  FNN     & 00m:54s     & 20h:43m:09s & 08m:46s    & 4.58ms     \\ \hline
\end{tabular}
}
\end{table}

\textbf{Results.} According to Table~\ref{tab:improv_evaluations}, the inclusion of \name{} with FNN surrogates contributed to significant decrease ratios in prediction error, from $12\%$ to $47\%$, while the baseline leads to a maximum of $1.25\%$ decrease rate on the GP’s prediction errors. Hence, \name{} was able to cover many problematic OOD inputs with high errors that lead to considerable reduction in FNN prediction errors. Besides, the speed up rates obtained by \name{} plus FNN surrogates are mostly higher than those obtained by Baseline plus GP counterparts. Indeed, \name{} was able to detect the underlying OODs more accurately with fewer false alarms than the baseline, which turns almost three hours of runtime to a few minutes. FNN surrogates have already better speed up in the inference time than their GP counterparts (see Table~\ref{tab:improv_evaluations}). Table~\ref{tab:surrogate_step_times} also reports lower values of training time, validation time and test time obtained by FNN surrogates compared to those yielded by GP counterparts. This can be explained by the FNN's deterministic mapping function that includes consecutive weighted sums and ReLU activations, and by its learning algorithm that applies loss gradients w.r.t. the parameters to iteratively update them. On the other hand, GP’s mapping function is stochastic including multivariate posterior gaussian distributions, and its approximate Bayesian learning that updates our prior belief in our gaussian parameters in line with marginal log-likelihood estimates on the observed data. Nonetheless, Table~\ref{tab:surrogate_step_times} shows that only FNN surrogates require a time-consuming and hand-crafted hyperparameters tuning to select the structure of the model (i.e., depth and width), configure the optimizer, adapt the regularizer strength, etc. All these choices are guided by trial-and-error processes on validation set. In contrast, the GP surrogate design is straightforward, and the kernel hyperparameters are determined systematically within the Bayesian posteriori optimization. The tuning of FNNs is common for any DL solution; so GPU-enabled parallelism and multi-machine distribution is straightforward when using modern DL frameworks.
\begin{table}[ht]
\centering
\caption{Computation Times for different steps of OOD Detection Method Design.}
\label{tab:ood_step_times}
\resizebox{220pt}{!}{%
\begin{tabular}{|c|c|c|c|c|}
\hline
Method            & Train\_Time & Tune\_Time & Valid\_Time & Test\_Time \\ \hline
\textbf{Baseline} & 0.32s       & 178.99s    & 3.17s       & 271.08ms   \\ \hline
\textbf{SmOOD}    & 0.43s       & 12.47s     & 4.22s       & 2.16ms     \\ \hline
\end{tabular}
}
\end{table}

Table~\ref{tab:ood_step_times} demonstrates that \name{} outperforms the baseline in terms of low design and evaluation workloads. The reason is the fast computation of predictions and derivatives using FNN surrogates to construct the local sensitivity profiles, and the ease with which the \name{} inherent classifier and oversampler can be trained and tuned due to the low dimensionality of the precomputing sensitivity profiles. This is contrary to the baseline which requires repeated distance calculations between a requested design configuration and all the validation set to identify the nearest neighbors.

\FrameSep.5em
\begin{framed}
\noindent
\textbf{Finding 4:} \name{} plus FNN surrogate enables accelerated and accurate hybrid optimization, achieving, on average, $34.65\%$ and $58.36\times$ of decrease error rate and computation speed up rate.
\end{framed}

\textbf{Domain Expert Feedback.}
According to domain experts, as long as the tuning process is automated, the FNN is still viable given the achieved error rate regardless of the build time. Furthermore, they emphasize that these FNN surrogates are mapping aircraft design variables to a particular aircraft performance attribute. Hence, an optimized FNN surrogate can be involved in many design optimization studies, which compensates for its relatively-heavy creation procedure. Most changes applied to the HF simulations are adjustments to the design variables in order to accommodate new requirements recommended by marketing analysts. This may result in further training on FNN parameters rather than structural/hyperparameters tuning. Creating FNN surrogates from scratch can be only done in response to less-frequent, major updates to HF model inner functions and capabilities. Nevertheless, experts suggest taking full advantage of the hybrid mode and storing the detected OOD inputs to further fine-tune the FNN after every optimization cycle since the training is relatively fast.
\section{Conclusion}
\label{sec:conclusion}
In this research work, we demonstrate that feedforward neural networks outperforms Gaussian processes across several surrogate models of aircraft design performance, especially when the predicted quantity is highly nonlinear. These high-capacity black box models are prone to out-of-distribution issues, which restricts their direct use in aircraft design optimization. Due to the risk of overconfident uncertainties against these OOD samples, SmOOD effectively spots them by identifying their suspicious local sensitivity behaviors that are far from the level of FNN smoothness observed across validation. Statistically, SmOOD reveals up to 93\% of OOD samples, and its use as router in hybrid surrogate aircraft design optimization leads to $34.65\%$ and $58.36\times$ of error reduction and runtime speed up rates. Our empirical evaluation reinforces our prior belief in the a priori smoothness that exists over the HF simulations data. Indeed, we expect similar priors to exist in many other applications, on which our findings can be extrapolated, and our local sensitivity profiling can serve as a surrogate for model uncertainty, and as a discriminatory criterion to separate ID from OOD. The next step will be to develop an active version of SmOOD that stores the revealed OODs to fine-tune the FNN and expand further the ID regions, which results in fewer requests to the HF model and faster design assessments.
\section*{Acknowledgment}
This work is supported by the DEEL project CRDPJ 537462-18 funded by the National Science and Engineering Research Council of Canada (NSERC) and the Consortium for Research and Innovation in Aerospace in Québec (CRIAQ), together with its industrial partners Thales Canada inc, Bell Textron Canada Limited, CAE inc and Bombardier inc. We also acknowledge the support of the Canadian Institute for Advanced Research (CIFAR).
\balance
\bibliographystyle{ACM-Reference-Format}
\bibliography{sample-base}
\end{document}